# Crossmodal Language Grounding in an Embodied Neurocognitive Model

*Stefan Heinrich*[1,2]*, *Yuan Yao*[3], *Tobias Hinz*[1], *Zhiyuan Liu*[3], *Thomas Hummel*[1], *Matthias Kerzel*[1], *Cornelius Weber*[1] and *Stefan Wermter*[1]

[1] Knowledge Technology Group, Department of Informatics, Universität Hamburg, Hamburg, Germany, [2] International Research Center for Neurointelligence, The University of Tokyo, Tokyo, Japan, [3] Natural Language Processing Lab, Department of Computer Science and Technology, Tsinghua University, Beijing, China

Human infants are able to acquire natural language seemingly easily at an early age. Their language learning seems to occur simultaneously with learning other cognitive functions as well as with playful interactions with the environment and caregivers. From a neuroscientific perspective, natural language is embodied, grounded in most, if not all, sensory and sensorimotor modalities, and acquired by means of crossmodal integration. However, characterizing the underlying mechanisms in the brain is difficult and explaining the grounding of language in crossmodal perception and action remains challenging. In this paper, we present a neurocognitive model for language grounding which reflects bio-inspired mechanisms such as an implicit adaptation of timescales as well as end-to-end multimodal abstraction. It addresses developmental robotic interaction and extends its learning capabilities using larger-scale knowledge-based data. In our scenario, we utilize the humanoid robot NICO in obtaining the EMIL data collection, in which the cognitive robot interacts with objects in a children's playground environment while receiving linguistic labels from a caregiver. The model analysis shows that crossmodally integrated representations are sufficient for acquiring language merely from sensory input through interaction with objects in an environment. The representations self-organize hierarchically and embed temporal and spatial information through composition and decomposition. This model can also provide the basis for further crossmodal integration of perceptually grounded cognitive representations.

***Correspondence:**
*Stefan Heinrich*
*heinrich@informatik.uni-hamburg.de*

## 1. INTRODUCTION

While research in natural language processing has advanced in specific disciplines such as parsing and classifying large amounts of text, human-computer communication is still a major challenge, due to multiple aspects: speech recognition is limited to good signal-to-noise conditions or well-adapted models, dialogue systems depend on a well-defined context, and language elements are difficult to reconcile with the environmental situation. Consequently, interactive robots that match human communication performance are not yet available. One reason for this is the fact that the crossmodal binding between language, actions, and visual events is not yet fully understood and was thus not realized in technical systems that have to interact with humans (Hagoort, 2017).

Imaging techniques such as Functional Magnetic Resonance Imaging (fMRI) have provided a better understanding of which areas in the cortex are involved in natural language processing and that these areas include somatosensory regions. Language studies have shown that there is a tight involvement of crossmodal sensation and action in speech processing and production as well as in language comprehension (Friederici and Singer, 2015). Thus, there is increasing evidence that human language is embodied. This means that it is grounded in most sensory and sensorimotor modalities and that the human brain architecture favors the acquisition of language by means of crossmodal integration (Pulvermüller, 2018).

As a consequence, research on cognitive modeling and developmental robotics is working toward developing models for natural language processing that reflect our understanding of distributed processing and embodied grounding of language in the brain. This way, the overall goal of studying the problem of language grounding in crossmodal perception and action can get approached. A particularly important aim is to develop a model for language grounding which reflects bio-inspired mechanisms and minimized difficult assumptions for the computational mechanisms.

In this paper, we present an embodied neurocognitive model for crossmodal language grounding that is trained in an end-to-end fashion. Additionally, we explore the concepts of varying multiple timescales in processing as well as distributed cell assemblies in representation learning. Based on the proposed model, we aim to investigate the characteristics of the learned crossmodally integrated representations.

## 1.1. Related Work

In order to bridge the gap between formal linguistics and bio-inspired systems, several valuable computational models have been developed that bring together language and an agent's multimodal perception and action. In their seminal Cross-channel Early Lexical Learning (CELL) model, Roy and Pentland (2002) demonstrate word learning from real sound and vision input. Each of these inputs is processed into a fixed-length vector, then lexical items arise by associations between vectors that represent the corresponding speech and an object's shape. Roy (2005) also highlights the importance of combining physical actions and speech in order to interpret words and basic speech acts in terms of schemas, which are grounded through a causal-predictive cycle of action and perception. Several works use self-organizing maps (SOMs), e.g., to form joint neural representations of simulated robot actions and abstract language input to encode the corresponding sensory-motor schemata (Wermter et al., 2005). This model addresses mirror neurons found in the motor cortical region F5, which link actor and observer by activating when performing a corresponding action or even just seeing or hearing it performed by someone else (Rizzolatti and Arbib, 1998). Vavrečka and Farkaš (2014) use a RecSOM (Voegtlin, 2002) which has a recurrent architecture with recursive updates to handle sequential input. Using a RecSOM and multiple SOMs, arranged in parallel for linguistic and visual input, and hierarchically for the integration of modalities, the model grounds spatial phrases within the corresponding image information.

Recent works often make reference to biological findings that support grounded language processing. Friederici and Singer (2015) provide evidence that linguistic and other cognitive functions are based on similar neuronal mechanisms, for example, single neurons react similarly to seeing a picture of a person's face or reading the person's name. More generally, Pulvermüller et al. (2014) propose a cognitive theory of distributed neuronal assemblies or thought circuits, integrating brain mechanisms of perception, action, language, attention, memory, decision, and conceptual thought. Rather than by SOMs, these neuroscience findings are better accounted for by distributed neural firing models. For example, in a multi-area model of cortical processing (Garagnani and Pulvermüller, 2016), some neurons become category-general while others are in category-specific semantic areas.

Among recurrent neural models, the multiple timescale recurrent neural network (MTRNN) (Yamashita and Tani, 2008) allows the emergence of a functional hierarchy with reusable sequence primitives. Heinrich and Wermter (2018) ground the generation of language in visual and motor proprioceptive signals, showing that an MTRNN can self-organize latent representations that feature hierarchical concept abstraction and concept decomposition. Zhong et al. (2019) address the generalization ability of MTRNNs by making use of semantic compositionality of simple verb-object sentences. They train an iCub robot to produce action sequences following a simple verb-object sentence comprising a selection of 9 verbs and 9 objects, where the network generalizes to all combinations despite being trained only on a subset. Yamada et al. (2017) investigate the handling of logic words in sentences from which an Long Short-Term Memory (LSTM) network generates corresponding robot actions. They show that, for example, the word "and" works like a universal quantifier, while the word "or" creates an unstable space in the LSTM dynamics.

While these models are used unidirectionally, bidirectional models have been proposed that can map both perceived language commands to actions and perceived actions to language descriptions. For this task, Yamada et al. (2018) train two paired recurrent autoencoders, one encoding the textual description sequence, the other encoding the action sequence. The autoencoders are paired by a joint loss function term that drives the two autoencoders' center-layer representations, which both have the same dimensionality, to be similar. As a result, a textual description leads to a representation that is suitable to generate an action sequence, and vice versa. For interactivity, the action sequence autoencoder receives additional image input in both encoder and decoder, while producing only the joint angle sequences as output. In each autoencoder, the direction of information flow between layers is fixed from input toward the output. In contrast, Antunes et al. (2018) implement a model of truly bidirectional information flow between three recurrent MTRNN layers of fast, medium, and slow timescale units. A subset of the fast units acts as input (or output) to a robot action sequence, and a subset of the slow layer's units acts as output (or input) to the language description. However, it needs to be investigated whether neural groups emerge that are solely devoted to information transmission into one of the directions, or, rather, whether shared bidirectional functionality emerges.

Another line of recent works shows that enriching linguistic data with other modalities can lead to better-performing systems. For example, continuous word representations like word2vec (Mikolov et al., 2013) or GloVe (Pennington et al., 2014) have become popular, since they span some semantically meaningful low-dimensional space leading to robustness and to the possibility to track relations between words. Additionally, the original words can be recovered from the representations even when they are corrupted or altered by noise. These embeddings can become even more powerful when involving multiple modalities. Hill and Korhonen (2014) train a word2vec-like model on the ESPGame dataset, which annotates images with a list of lexical concepts, and on the CSLB Property Norms dataset which contains concepts for which human annotators produced several semantic properties. Lazaridou et al. (2015) train a similar model on text from Wikipedia and add visual information from the ImageNet database to a subset of the words, which is processed into an abstract vector by a pre-trained Convolutional Neural Network (CNN). Wang et al. (2018b) use GloVe vectors pre-trained on the Common Crawl dataset together with CNN-based visual vectors pre-trained on ImageNet. Auditory features extracted from a CNN network pre-trained on Google's AudioSet data are included in Wang et al. (2018a). The results of these models show that multimodal embeddings outperform unimodal embeddings. Furthermore, suitable images can be generated not only for concrete words but also for some abstract words by selecting the nearest neighbor image for a generated image vector (Wang et al., 2018a). For reinforcement learning interactive game agents, it was shown that augmenting environmental information with language descriptions (Narasimhan et al., 2018) or instructions (Chaplot et al., 2018) leads to better generalization and transfer capabilities.

There is also a recent focus on tasks like image captioning, Visual Question Answering (VQA), and phrase grounding in images. In these tasks, sequentially processed language refers to elements of images and the availability of corresponding large datasets for supervised learning has driven model development. VQA research, for example, led to neural architectures that facilitate reasoning steps, e.g. by affine transformations within the visual processing stream based on conditioning information from the question (Perez et al., 2018), by novel recurrent Memory, Attention, and Composition (MAC) cells (Hudson and Manning, 2018), or by more explicitly using graphs for reasoning (Hudson and Manning, 2019). Yet, these models do not cover the production of language, since VQA tasks are cast as classification problems where the network produces only the label to the correct answer among a given set of answers. Instead, they are tailored toward reasoning, but often fail in generalization, if their architecture is not primed for the task (Santoro et al., 2017). A potential reason for the lack of generalization can be in the poor integration of language and image representations by these models, since they are not embodied in interactive agents, which Burgard et al. (2017) suggest.

Overall this shows the need for an embodied neurocognitive model that can help to explain language processing in the brain and at the same time proves to be effective in generalization. To this end, we need to more closely look into components of both temporal decomposition and composition and at the same time realize an inherent multimodal abstraction on both sensory as well as conceptual level. It seems crucial that temporal decomposition and composition directly emerges in a model based on the context or the data, while multimodal abstraction needs to take place on sensory up to an overall contextual level.

### 1.2. Contribution

In this paper, we develop a neurocognitive model that grounds language production into embodied crossmodal perception. In particular, our model aims to map the auditory, sensorimotor, and visual perceptions onto the production of verbal utterances during the interaction of a learner with objects in its environment.

As a core characteristic, the model allows for the implicit adaptation of timescales based on the temporal characteristics of both perception and language production. Furthermore, the model tests multimodal abstraction in an end-to-end fashion with limited constraints on the preprocessing of the sensory input. The model is analyzed in depth based on a developmental robotics data recording that mimics natural interactions of an infant with said objects. This Embodied Multimodal Interaction in Language learning (EMIL) data collection challenges the model by introducing a wider range of variability of the temporally dynamic sensory features, in order to exhibit effects on language learning and latent representation formation concerning findings for the human brain.

Therefore, the contribution of this paper is three-fold[1]:

- We present a neurocognitive model for language grounding which reflects bio-inspired mechanisms such as an implicit adaptation of timescales as well as end-to-end multimodal abstraction. It addresses developmental robotic interaction and extends its learning capabilities using larger-scale knowledge-based data.
- We demonstrate the effectiveness of our model on the novel EMIL data collection, in which the cognitive robot interacts with objects in a children's playground environment while receiving linguistic labels from a caregiver.
- We conduct an in-depth analysis of the model on the real-world multimodal data and draw several important conclusions. For example, crossmodally integrated representations are sufficient for acquiring language merely from sensory input through interaction with objects in an environment.

## 2. EMBODIED NEUROCOGNITIVE MODEL

In order to add insight to related computational models, we aim to develop a model that satisfies a number of constraints. First, we seek to minimize difficult assumptions for computational mechanisms. In particular, we avoid building on top of mechanisms that are appealing for machine learning but not yet proven or not plausible for the processing in the brain such as neural gating, dropout regularization, or residual connections. In fact, we aim at building on top of the most simple computational architecture that still allows studying our proposed mechanisms. Second, we work with a minimal level of assumptions regarding language grounding. Here, we avoid using an oversimplified language such as modeling on word-level only. Additionally, we do not use natural speech but rather a simpler phonetic representation as the desired output. We will build our computational model with a distinct focus on the following biological mechanisms.

[1]The source code of the model and experiment details can be found on https://github.com/heinrichst/adaptive-mtrnn-grounding.git.

## 2.1. Biological Inspiration

It has been suggested that the human cognition is particularly strong because the human brain is good at both information composition and decomposition (Murray et al., 2014). Furthermore, it seems that many processes in the brain are reused in or coupled to a range of cognitive functions. In the brain, the decomposition and composition are governed by neural oscillations, multiple timescales in hierarchical processing streams, and a complex interplay of neural populations and local integration by mode coupling (Buzsáki and Draguhn, 2004; Badre et al., 2010; Engel et al., 2013). Additional evidence suggests that in higher stages of the spatial or temporal hierarchy neurons are organized in cell assemblies (Damasio, 1989; Palm, 1990; Levelt, 2001). These sparsely connected webs of neurons are distributed over different cortical areas and both hemispheres and form consistently during development for concepts on higher or lower levels.

In language grounding, both multiple timescales and cell assemblies seem to be reused. Multiple timescales in processing have been reported across the brain from lower auditory processing up to higher processing of perception (Ulanovsky et al., 2004; Smith and Kohn, 2008; Himberger et al., 2018) and cell assemblies are suggested to activate for both word processing as well as the overall thought processes (van der Velde, 2015; Tomasello et al., 2019). As a consequence, in our computational model, we further study the mechanisms of multiple timescales in information processing as well as crossmodal fusion by and sequence activation from cell assemblies.

## 2.2. Computational Model

We base our computational model on the Continuous Time Recurrent Neural Networks (CTRNN) architecture because of its universality in modeling sequential signals. Although we can derive the CTRNN from the leaky integrate-and-fire model and thus from a simplification of the Hodgkin-Huxley model from 1952, the network architecture was suggested independently by Hopfield and Tank (1986) as a nonlinear graded-response neural network and by Doya and Yoshizawa (1989) as an adaptive neural oscillator. Specifically, the CTRNN can be understood as a generalization of the Hopfield Network (Hopfield, 1982) with continuous firing rates and arbitrary leakage in terms of time constants. Compared to the Simple Recurrent Network (SRN, or Elman Network), the timescale $\tau$ is an additional hyperparameter of asymptotically *not* leaking, thus, a neuron can maintain part of its information for a longer period of time.

The activation $\mathbf{y}$ of CTRNN units is defined as follows:

$$\mathbf{y}_t = f\left(\mathbf{z}_t\right), \tag{1}$$

$$\mathbf{z}_t = \left(1 - \frac{\Delta t}{\tau}\right)\mathbf{z}_{t-\Delta t} + \frac{\Delta t}{\tau}\left(\mathbf{W}\mathbf{x} + \mathbf{V}\mathbf{y}_{t-\Delta t} + \mathbf{b}\right), \tag{2}$$

for inputs $\mathbf{x}$, previous internal states $\mathbf{z}_{t-\Delta t}$, input weights $\mathbf{W}$, recurrent weights $\mathbf{V}$, bias $\mathbf{b}$, and an activation function $f$. The *timescale* can be a pre-determined common parameter $\tau$ for all neurons or a vector $\boldsymbol{\tau}$ of individual constants. In tasks with discrete numbers of time steps, the CTRNN can be employed as a discrete model, e.g., by setting $\Delta t = 1$.

With respect to modeling multiple timescales in information processing, the timescale parameter $\boldsymbol{\tau}$ provides an interesting mechanism to capture sequential aspects on different timescales or periodicities and is particularly crucial for the hierarchical abstraction capability of the Multiple Timescale Recurrent Neural Network (MTRNN, compare Yamashita and Tani, 2008). Our model, therefore, integrates this predefined hierarchical abstraction. In particular, a fixed number of layers is defined a priori, e.g., having three adjacent layers called *Input-Output* (IO, $\boldsymbol{\tau} = 2$), *Context-fast* (Cf, $\boldsymbol{\tau} = 5$), and *Context-slow* (Cs, $\boldsymbol{\tau} = 70$), in order to force the architecture to hierarchically compose or decompose information.

In order to achieve decomposition and composition in the MTRNN, the overall context of a sequence is learned by or stored into some of the units in the slowest layers, called *Context-controlling* (Csc) units. Consequently, such an MTRNN can be defined in two forms, providing a decoder and an encoder component.

- MTRNN with Context Bias: the Csc units operate as a parametric bias during *production* and thus the Csc values are learned backwards during gradient descent training (compare Awano et al., 2010). Since the network weights are trained in parallel to the Csc units, the MTRNN with context bias learns to decompose a temporally dynamic sequence from a static initial bias.
- MTRNN with Context Abstraction: the Csc units operate as abstracting a static output during *sensory processing* similar to one-point classification (compare Heinrich and Wermter, 2018). Due to the increasingly larger timescales in the layers, the network learns to compose a static overall context from a temporally dynamic sequence.

When an MTRNN with context bias is coupled with an MTRNN with context abstraction in an end-to-end architecture, the Csc values of both networks are updated iteratively and form latent representations similar to a sparse auto-encoder on sequences.

In the MTRNNs, however, the $\boldsymbol{\tau}$ needs to be carefully chosen as a hyperparameter, based on a priori known temporal characteristics of the data. This is usually done in coarse approximation on layer or module level. In contrast, time constants in the brain are subject to change during development and are hypothesized to be directly related to temporal structures (He, 2014). In previous work we developed a mechanism to obtain an *adaptive* timescale $\boldsymbol{\tau}^{\mathrm{A}}$ for each unit (Heinrich et al., 2018a). The timescales are governed by learnable weights **U** that work like a bias on the timescale instead of on the activation:

$$\boldsymbol{\tau}_t = \boldsymbol{\tau}_t^{\mathrm{A}} = 1 + e^{\mathbf{U}+\boldsymbol{\tau}_0}, \tag{3}$$

where the exponential function ensures timescales in $(1, \infty)$, and the vector $\boldsymbol{\tau}_0$ can be initialized with sensible values for the timescales while the weights **U** get initialized to values close to zero. As a rule of thumb, we can initialize $\boldsymbol{\tau}_0$ either at random between 1 and a reasonably large number, i.e., to the length of the expected longest sequence (or a logarithm thereof) (Heinrich et al., 2015), or with timescales that are known to work well for MTRNNs in similar tasks.

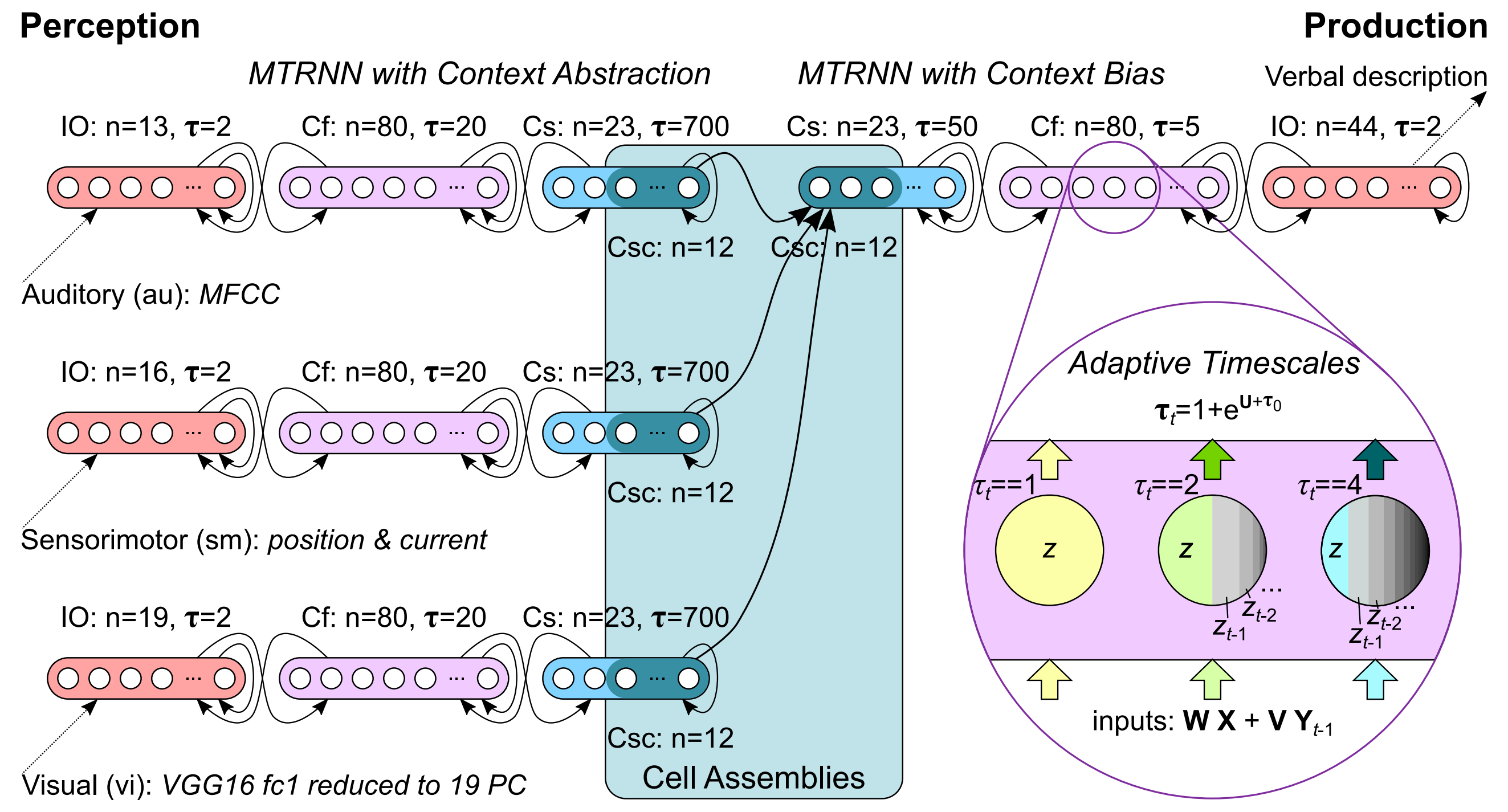

**FIGURE 1 |** Computational model: Adaptive MTRNNs with context abstraction for each input modality are coupled with an adaptive MTRNN with context bias via cell assemblies. Example timescales visualize the logarithmic leakage of information in the neurons.

In our computational model we, therefore, utilize adaptive MTRNNs with context abstraction for sensory inputs from multiple modalities and an adaptive MTRNN with context bias for verbalizing the observed sensation in natural language. Through this, the architecture provides a composition of a sensation into an overall meaning for that sensation as well as a decomposition of a meaning into a verbal description. The Csc units of all MTRNNs are coupled in cell assemblies from which, supposedly, a sparse latent representation for the meaning can emerge through iterative learning. Specifically, we integrate up to three MTRNNs for the abstraction of temporal dynamic auditory (au), sensorimotor (sm), and visual (vi) perception as well as an MTRNN which uses this context for language production in terms of verbal utterances describing the perception. The overall architecture is illustrated in **Figure 1**, further details on the scenario are provided in section 3.

## 2.3. Developmental Robot Scenario for Language Grounding

To investigate language grounding, we couple multi-modal sensations and a verbal description in order to train our model in an end-to-end fashion. Although supervised, this is related to models that investigate language grounding by mapping perception and action through Hebbian learning and studying the emergence and consolidation of connection patterns (e.g., Garagnani and Pulvermüller, 2016). Our aim is to further scale to a temporally dynamic scenario from real-word observations with the aim of studying both the emergence of timescales as well as connection patterns in terms of cell assemblies.

For this, our set-up is borrowed from a developmental robot scenario, where a humanoid robot, such as the Neuro-Inspired COmpanion (NICO, Kerzel et al., 2017), represents an infant learner who explores the environment by interacting with objects on a table and perceives verbal descriptions from a caregiver for particular object manipulations (see **Figure 2**). We conducted a data collection of the EMIL data set[2] (Heinrich et al., 2018b), that includes parallel multi-modal recordings from the robot's body-rational view as well as visual observations from a teacher perspective. The robot performs an action from a set of four predefined motions on a set of 30 distinct objects which exhibit different shape, color, texture, weight, friction, and sound characteristics when moved. The interaction is captured by microphones in the robot's ears for 48 kHz auditory sensation, by proprioception in the arm (motor position and current from eight motors, with 30 read-outs per second) for sensorimotor perception, and by a 90 degree field-of-view and 30 fps camera for visual perception. In addition, a textual description was recorded that describes the interaction with the object.

[2]More details on the collection are provided in the **Appendix**. We plan to obtain several versions of the EMIL data set with increasing scenario complexity and amount of data. The version 1 is publically available via https://www.inf.uni-hamburg.de/en/inst/ab/wtm/research/corpora.html.

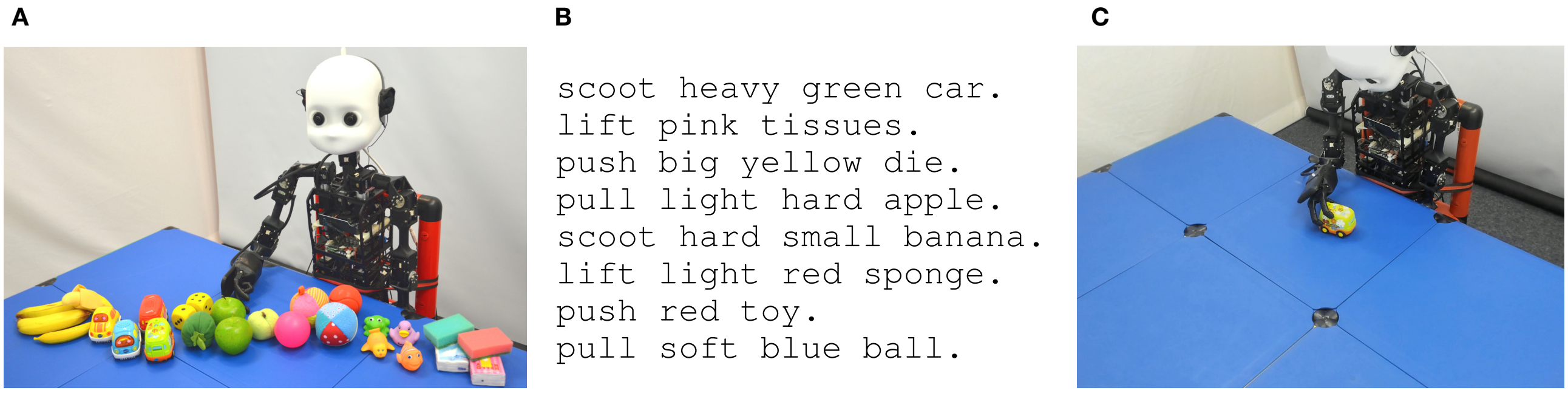

**FIGURE 2 |** Developmental robot scenario of the EMIL data collection (Heinrich et al., 2018b): NICO is interacting with objects and perceives the interaction on auditory, sensorimotor, and visual modalities. A teacher provides a description for the interaction. **(A)** Scenario, **(B)** Example descriptions, **(C)** Teacher perspective.

To study the model on this scenario, we prepared two data sets from the EMIL version 1 collection:

- EMILv1 Data: 240 sensation-description pairs with up to 740 time steps for the perception streams and a simple holo-phrase with up to four words for the description. The descriptions were created from a vocabulary of 68 words and 4 symbols for punctuation, where a word is represented with one to nine phonemes.
- EMILv1 + Teacher Data: in order to mimic the situation of a caregiver providing additional descriptions to foster the infant's learning, we extended the data with additional teacher input. In particular, we appended data points where we replaced the nouns and verbs with synonyms and added slight Gaussian noise to the perception ($\sigma = 0.01$) in order to obtain 2,880 unique pairs. This is motivated by infants learning language better through scaffolding and guidance from their parents (Tomasello, 2003). The process can also be viewed as data augmentation from linguistic knowledge, which results in increased diversity and scale of crossmodal data for language learning, and is shown to lead to better generalization ability of neural models (Zhang et al., 2015). In order to ensure the quality of the teacher data, synonyms are obtained from WordNet (Miller, 1995), a high-quality lexical knowledge base according to the sense of the replaced word.

The EMILv1 data exhibits a couple of interesting characteristics. On the one hand, with the particularly long and noisy sequences (especially in the sensorimotor modality) the training is challenging for RNNs. On the other hand, in most sequences, the visual modality is most informative for the presented action + object pair. Compared to previous developmental robotic data sets, e.g. in Heinrich and Wermter (2018) the data does not imply a necessity for superadditivity (i.e., that more information is gained from multiple modalities only) but rather selectivity (meaning that one modality might be strongly favored in certain situations).

## 2.4. Representation and Training

For the verbal descriptions we prepared two different language representations:

- Phonetic: we transformed the utterances into phonetic sequences based on the ARPAbet and dictionary provided by CMU[3] and represented these sequences as simple one-hot vectors. This is different from previous related research (Hinoshita et al., 2011; Heinrich and Wermter, 2018) where a single phoneme was stretched backwards and forward in time and thus learned much easier by using teacher forcing.
- Word embedding: in order to study the model on both fine-grained phonetic-level and coarse-grained word-level we utilize the GloVe-6B embeddings provided by the Stanford NLP group (Pennington et al., 2014).

We expect that the phonetic representation is more challenging and provides the necessity for the emergence of temporal composition in the MTRNN for verbal descriptions. The word embeddings, on the other hand, are more informative for studying the multi-modal fusion since the word embeddings already reflect semantic relatedness.

For the multi-modal sensation, we perform some simple preprocessing in order to provide input streams of comparable dimensions and low-level feature abstraction. For the auditory input, we transform the signals using Mel-Frequency Cepstral Coefficients (MFCC) analysis into 13 dimensions with a frame size of 33 ms and input window 60 ms. This is acceptable in terms of biological inspiration as the cochlea is doing a Fourier transformation of auditory signals that are roughly similar. The sensorimotor input was taken as is, but normalized, to result in 16 dimensions. The visual input in terms of a video stream was processed by a VGG16 neural network (Simonyan and Zisserman, 2015) (we took the output of the first dense layer after the convolution and pooling layers) and further condensed to 19 dimensions by Principal Component Analysis (PCA) in order to provide visual features. The VGG architecture was chosen since it is a powerful CNN architecture that was developed based on biological inspiration but does not yet incorporate implausible mechanisms such as arbitrary residual connections (Krüger et al., 2013; Hu et al., 2019). In our model, we used VGG layers that were pre-trained on ImageNet and thus provide reasonable features for objects. The reduction with PCA is not supposed to mimic any specific cortical processing but is an easy step in systematically reducing complexity in the model, which alternatively could be realized by neural unsupervised learning as well.

[3] ARPAbet is an American English phonetic transcription set, transcribed in ASCII symbols, http://www.speech.cs.cmu.edu/cgi-bin/cmudict.

Since all network parameters are fully differentiable (Heinrich et al., 2018a), the architecture can be trained end-to-end using gradient descent. Although for the brain theories are in favor of Hebbian learning during development instead of backpropagation, we argue that for our research aim of studying the emergence of multiple timescales and the emergence of crossmodally fused representations for language grounding a supervised error signal is feasible (Dayan and Abbott, 2005; Lillicrap and Santoro, 2019).

## 3. EVALUATION AND ANALYSIS

In order to analyse our model for how compositional language is grounded in multimodal sensations and how multimodal abstraction emerges through learning, we trained different variants of our model on different variants on the EMIL data sets.

For all experiments, we optimized the hyperparameters, i.e., the architecture size, optimization algorithm, learning rate, and batch size. We started with the model architecture from baseline CTRNNs, which are configured with equal timescales $\tau = 1$ for all neurons. Once good hyperparameters were found, we used the same hyperparameters for all MTRNNs while separately optimizing their timescales. These timescales, in turn, are used as initial timescale values of the adaptive MTRNNs (AMTRNNs). All models were trained for at most 5,000 epochs and a validation set was used for early stopping. We performed a 10-random sub-sampling validation, i.e., we repeated each run ten times with a different and independent split of training, test, and validation data (75, 12.5, 12.5%) as well as different and independent weights-initialization, based on a different random seed. The best results were found with RMSprop (Tieleman and Hinton, 2012), a learning rate of 0.01, and a batch size of 30. The exact architectural parameters are noted in **Figure 1**. In the following, for the argmax on the output, we report the mean accuracy over the cross-validation for each setup.

### 3.1. Generalization on Developmental Interaction Data

As a first step, we are interested in how well the architecture can actually learn verbal descriptions for the different sequential inputs. In order to inspect the generalization, we compare the accuracy on the test sets for both data sets, both verbal utterance representations, and three different model variants. In particular, we compare the baseline CTRNNs with the optimized MTRNNs and AMTRNNs.

The accuracy results (including standard errors) are presented in **Table 1**. We observe that the generalization is difficult for all models and that utterances which were described entirely correct are rare. For the phonetic representation, the model produces descriptions with a range of small errors such as pauses that are too long or producing incorrect phonemes at the end of words (rare) or of the utterance (more common). In many of those cases, the model shows tendencies to produce wrong descriptions from the first incorrect phoneme onward. For the word embedding representation, the descriptions are overall better, but in some cases, words are mixed up that are not necessarily semantically related.

Nevertheless, we observe strong differences between the models with different timescale characteristics on both the EMILv1 data and the data extended with additional teacher input (significant different performance between baseline CTRNNs and both other models, with $p < 0.05$). The baseline CTRNN model is not able to derive any description completely correct for the phonetic representation. In fact, we found that the CTRNN fails after the first few phonemes and afterwards just produces the phoneme that is most common in the data (usually the pause symbol *SIL*). For the word embedding, the performance is better, indicating that the CTRNN can handle the short utterances describing the sequence (only up to five words, compared to up to 25 phonemes in the phonetic representation). This also means that the CTRNN is able to capture the meaning of the input sequences (with up to 740 time steps) in terms of the presented *action + object*. The model based on an MTRNN with optimized timescales shows a large improvement on the phonetic representation. The model using adaptive MTRNNs performs even better (but not significant, with $p > 0.05$). Here, the errors in production are distributed over the utterance and a mostly incorrect description is characterized by the production of semantically wrong words, although the words were spelt correctly. Both the MTRNN- and AMTRNN-based models show improvements on the word embedding representation but notably differ in their mistakes. The incorrect words for the CTRNN seem arbitrary, especially if the words are at the end of the utterance. For the MTRNN and AMTRNN, we notice that incorrectly produced words were in many cases semantically related, e.g., mixing up “light” with “hard” or “red” and “pink.”

Overall it seems that the correct description is strongly dependent on whether the latent distributed representation (the cell assemblies) in the Csc units is able to abstract the sensory input and, thus, if the composition in the sensory CTRNN/MTRNN/AMTRNN correctly captures the temporally distributed information. In the following, we will, therefore, analyse the temporal aspect as well as the latent representations.

### 3.2. The Role of Adaptive Timescales

In order to inspect how the individual timescales contribute to sensory abstraction and utterance production, we compare the developed timescales as well as the activations within the AMTRNNs during processing the data. In **Figure 3**, we show a representative example for an interaction labeled "scoot heavy green car." This sample is not producing the description (entirely) correct but shows characteristics that we found regularly in many cases. In **Figures 3A–C**, we compare the neural activation in all neurons with the raw input data, for auditory input shown as a spectrogram in the frequency domain, for sensorimotor as the plain measurements, and for visual as selected frames during the interaction (**Figure 3F**).

**TABLE 1 |** Test accuracy (%) for different CTRNN architectures on phonetic vs. word representation.

| Model characteristic | EMILv1 data | | EMILv1 + Teacher data | |
|---|---|---|---|---|
| | **Phonetic** | **Word embedding** | **Phonetic** | **Word embedding** |
| Baseline CTRNNs | 25.472 ± 0.765 | 56.115 ± 2.412 | 18.476 ± 0.118 | 37.991 ± 0.226 |
| Optimized MTRNNs | 42.087 ± 0.868 | 63.309 ± 1.260 | 34.655 ± 0.418 | 51.896 ± 1.604 |
| AMTRNNs | 43.327 ± 1.025 | 64.029 ± 1.975 | 35.506 ± 0.461 | 54.691 ± 0.502 |

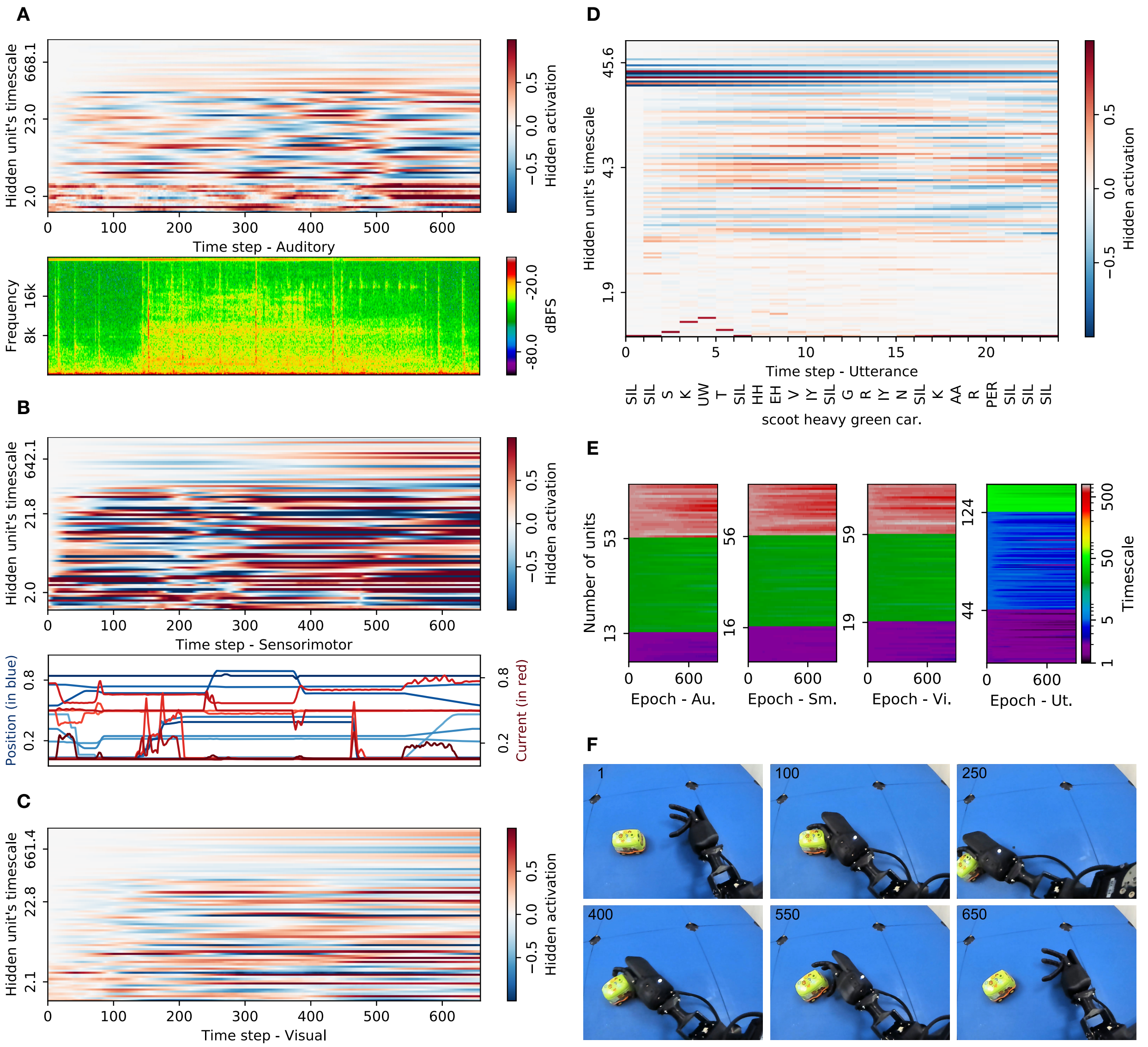

**FIGURE 3 |** Impact of adaptive timescales in processing crossmodal input and phonetic output sequences on a representative example: "`scoot heavy green car`." Hidden activations of all AMTRNN layers (stacked for each modality and sorted by timescale value) are shown together with the respective input or production. For the visual input, six frames are shown for selected time steps. **(A)** Auditory adaptive MTRNN and input. **(B)** Sensorimotor adaptive MTRNN and input. **(C)** Visual adaptive MTRNN. **(D)** Phonetic production adaptive MTRNN. **(E)** Timescale development during training. **(F)** Visual input (exemplary frames).

For both sensorimotor and visual activation we observe an increasing activity in the neurons with the highest timescales (in the graphs around a timescale of 660), showing that information is accumulated for the neurons that are part of the cell assemblies. For the auditory activation, this occurs on a much weaker level. We can also see that in the sensorimotor activation, neurons activate after some remarkable events, such as the spikes in the motor current around the first and second third of the sequence. This shows that, across the spectrum of timescales, neurons begin to reverberate when the current input seems different from sensory input in other interactions. Interestingly, in both sensorimotor and visual activations, neurons on timescales between 4 and 25 maintain their activation until the end of the sequence once positively or negatively activated. For the auditory activations, we can not easily spot a similar behavior but rather observe strong fluctuations for the neurons with small timescales until 80% of the sequence. Semantically plausible reverberations are rare, thus it seems the auditory information is much noisier and less decisive compared to the other modalities.

In the production of verbal utterances (**Figure 3D**) we spot patterns that are typical for MTRNNs: some neurons on lower timescale fluctuate according to specific phonetic output and neurons around timescales $4 - 6$ activate and maintain their activation for some time spans. In notable cases, these activations coincide with the production of words representing semantically meaningful phoneme chains. The neurons with lower timescales of around 42, however, keep their activations over time with some leakage. These timescales correspond to the IO, Cf, and Cs layers and indicate a hierarchical decomposition. Notable is that the correspondence of activity in the Cf layer, with a produced word, is less pronounced than expected, while the activations of specific phonemes fade quickly. Correct phonemes are still produced, but at some point only *SIL*s are activated. This clearly shows that this model has not ideally learned the production of the utterance, although the network structure induces the mentioned decomposition.

Regarding the learning of individual timescales, we see in **Figure 3E** that all AMTRNNs tend toward more fine-grained timescales in all layers. For the sensory input AMTRNNs, these changes are most notable for the neurons in the Cs layers, as they tend to result in smaller timescales (around 650) instead of the layer-wise optimized value of 700 of the MTRNN model. For the production AMTRNN, individual timescales also result in smaller values in some cases and a strong differentiation of the neurons in all layers. This indicates that, in addition to the predefined hierarchical structure, the AMTRNNs further adapted to the specific scales of relevant events in the sequences.

Overall it is notable that the timescale mechanism, w.r.t. the leakage of information, has its limit for covering events that occur on different timescales but are not particularly regular. In many cases, the multi-sensory perception is abstracted in terms of neurons accumulating information relatively independent of the timescales. The input data from the EMIL data set does not consist of chains of events that need to be composed, but they do show key events, such as grasping the objects or perceiving a difference in inertia through different current values in cases of rapidly moving an object. These key events seem to be captured, but neurons activate as a memory rather than a shortly active detector of features on a mid-level timescale. The production of verbal utterances, in many cases, illustrates shortcomings toward the end of the utterances, with the tendency of producing the overall most frequent phoneme (*SIL*).

## 3.3. Latent Representations in Cell Assemblies

Finally, we are interested in how cell assemblies form, based on the sensory input and description output. Specifically, we aim to inspect whether latent representations in the Csc spaces reflect the meaning of the utterances. We hypothesize that in cases of "good" models, the semantic components (action and object characteristics) that are exactly identical (e.g., the same action) or similar (e.g., a rectangular toy shape and a rectangular tissue shape) are represented similarly as well.

To analyse this, we compare setups where we trained AMTRNNs with all three modalities (auditory, sensorimotor, and visual), combinations of two modalities, or only on a single modality as input. The overview of the performance (accuracy results and standard errors) for these setups is presented in **Table 2**. For the trained networks we obtained the neural activations of the Csc units for the respective input AMTRNN and verbal description output AMTRNN and reduced the dimensionality of the representation to two Principal Components (PC) using PCA. For typical results and selected combinations of modalities, the reduced representations are plotted in **Figure 4**. Since the Csc from the sensory inputs map to the Csc for the verbal description we would expect that the plots for the verbal utterances show similarities most clearly. Note, however, that although two PCs usually explain $> 60\%$ of the variability, they are only one perspective on the representation among others. Nevertheless, we selected cases that are representative for our observations across the results and avoided using t-Distributed Stochastic Neighbor Embedding (t-SNE) instead of PCA in order to not introduce additional biases.

Surprisingly, the results indicate that the setup that only uses visual input data performs best, compared to setups that process multimodal input data (notable but not significant, with $p > 0.05$). Overall, the setups that have access to the visual modality perform better (significant for all combinations, with $p < 0.05$), whereas the auditory modality leads to worse results (significant for combinations with an auditory input vs a visual input, with $p < 0.05$). When inspecting representations of the cell assemblies we can identify an explanation in the emerging representations. The semantic components are best distributed in the visual modality, indicating clusters for most of the characteristics, e.g., the object shape and action. To see this, compare all panels for the visual modality in **Figure 4**. Even though we do not visualize this here, we found similar clusterings for the color semantic component. In the sensorimotor modality the clusters are particularly obvious for action but strongly overlap for the shape component (not shown: it also overlaps for color components as well as weight and softness). In the auditory modality, all semantic components overlap for the case of full multimodal input (**Figure 4A**) and unimodal input (**Figure 4D**). However, in case of the auditory representation being presented together with sensorimotor or visual information only, we found a slight tendency of clustering toward the clusters that emerged within the other input modality (compare **Figure 4B** for auditory and sensorimotor and **Figure 4C** for auditory and visual). In most cases, the representation in the Csc of the verbal utterance production showed a mixture of the representations in the input Csc.

**TABLE 2 |** Test accuracy (%) for training on restricted sensory input.

| Sensory input | au + sm + vi | au + sm | au + vi | sm + vi |
|---|---|---|---|---|
| EMILv1 data | 43.327 ± 1.025 | 35.709 ± 1.004 | 41.831 ± 0.958 | 44.252 ± 0.979 |
| EMILv1 + Teacher | 35.506 ± 0.461 | 33.672 ± 0.540 | 34.974 ± 0.376 | 34.557 ± 0.326 |
| **Sensory input** | **au** | **sm** | **vi** | |
| EMILv1 data | 35.945 ± 0.895 | 38.957 ± 0.695 | 44.409 ± 1.097 | |
| EMILv1 + Teacher | 31.623 ± 0.439 | 29.734 ± 0.412 | 33.815 ± 0.455 | |

Overall it seems that (a) the characteristics of the raw data have a large influence, and (b) the end-to-end learning slightly favors a merging of the input modalities that is not directly beneficial. For (a), inspecting the raw data confirms our observation and expectation. In our raw data, we observe that the input streams are usually both quite noisy but also distinctive for some aspects. For example, the proprioception information from the motors (motor current) shows large deviations but for the human inspector it is easy to discriminate the different actions, while distinguishing between heavy and light objects (stronger vs. lower current) or hard and soft objects (stronger squishing and thus different finger motions) is extremely hard. In the auditory recordings, it is not possible to discriminate most object characteristics except for different friction sounds of heavy and light objects. However, distinguishing the actions by the motor sound is sometimes possible. For (b), we hypothesize that the amount of data in the EMILv1 data set is insufficient w.r.t. the complexity of the architecture, whereas the larger number of examples in the EMILv1 + Teacher set leads to a slightly different convergence. When comparing both data sets in **Table 2** we find a tendency of modality selection for the smaller data set and a tendency of superadditivity for the larger one.

## 4. DISCUSSION

In this paper, we investigated an embodied neurocognitive model to better understand the effects of adaptive multiple timescales as well as multi-sensory fusion mechanisms in grounding a temporal dynamic verbal description into temporal dynamic perceptions. For the model, we adopt that the human brain is reusing composition and decomposition as well as multiple sensory modalities in grounding natural language (compare section 2.1). Furthermore, in the model, we realize the merging of senses in a higher stage and inherently assume that the multiple timescales are in fact necessary (compare section 2.2). In our results, we found that adaptive timescales help in abstracting the information from temporally long and complex perceptions. Preparing the layers in these AMTRNNs with context abstractions toward an implicit hierarchy of multiple timescales forces a composition of an overall meaning from the crossmodal perception.

However, the concept of leakage in the AMTRNN specifically and in the MTRNN generally seems to reach its limit here. In previous studies, sequences were usually limited to $< 50$ time steps and, as a consequence, easily learned. In our experiments, perception inputs have $\approx 700$ time steps for which MTRNNs hardly converge, even if a large hierarchy of carefully optimized timescales is tested. Consequently, meaningful abstractions emerge to some extent but compared to other mechanisms in machine learning, like gating or time-windowed CNNs, the resulting representations and performance are limited (Chang et al., 2017). Thus, although the decomposition through neural processes, which operate on different timescales, seems to contribute to the human abilities of language grounding, it does not explain how we cope with the complexity of our daily sensory input.

We also found that using end-to-end learning cell assemblies, i.e., pairs of temporally static abstracted modal information and production biases, show a tendency to organize w.r.t. similarities of the semantic components (i.e., an action, object shape, object softness, and so on). This is in line with previous studies and general observations on gradient descent machine learning. However, for our more natural and noisy interaction data, it shows that a choice between superadditivity and modality-specificity does not necessarily simply emerge but might involve additional cognitive processes.

In the past, language acquisition and grounding models were usually tested on synthetic toy examples or very constrained and carefully designed scenarios (Cangelosi and Schlesinger, 2015). Crucially, aspects of language were omitted or robotic interactions were designed particularly systematic. In contrast, our current study uses the EMIL data collection which challenges the model by introducing a wide range of variability in terms of sensory noise, object characteristics, and skewed distributions thereof. It seems, however, that by reducing these constraints and capturing truly multimodal and natural interaction scenarios we can reveal novel, potentially incompatible, effects.

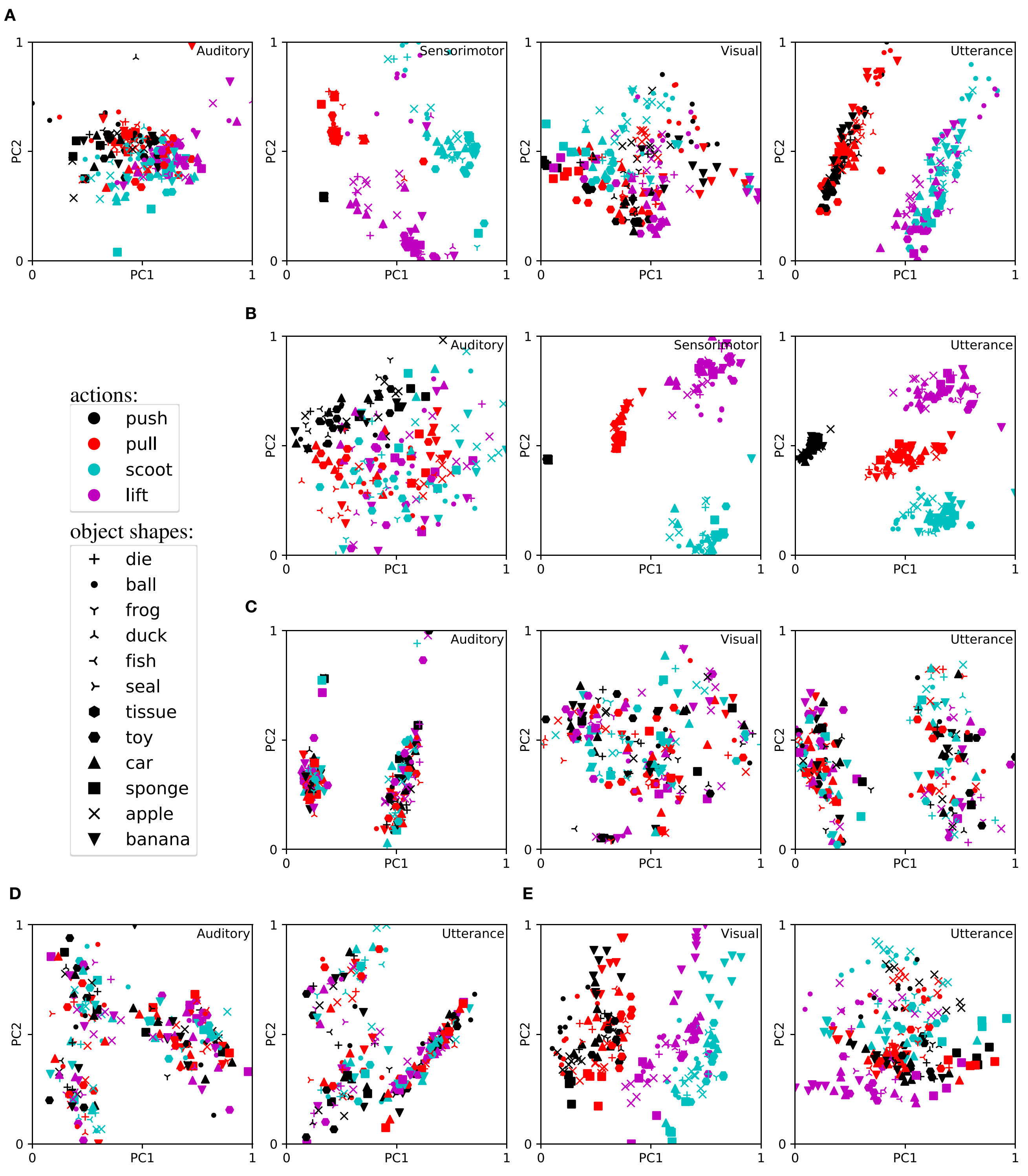

**FIGURE 4 |** Learned representations in the cell assemblies for training on different modalities (reduced by PCA to first the two principal components PC1 and PC2). **(A)** Perception via auditory, sensorimotor, and visual modalities. **(B)** Perception via auditory and sensorimotor modalities. **(C)** Perception via auditory and visual modalities. **(D)** Perception via auditory modality. **(E)** Perception via visual modality.

## 5. CONCLUSIONS

Overall, our embodied neurocognitive model shows that in an end-to-end learning architecture with hierarchical concept abstraction and concept decomposition, language grounding can emerge and generalize. Adaptive multiple timescales and multi-sensory fusion on concept level are, among others, effective components. Of similar importance are the scenario characteristics of our more complex and natural EMIL data collection, which introduces a larger range of variability and noise. Through using more complex data we observe novel effects such as limits in temporal abstraction and contradicting observations concerning superadditivity vs. modality-specificity.

For future research, when aiming to explain complex cognitive functions, we need to take into account the full complexity of the environmental context as well as of the computational conditions. For language acquisition and grounding it seems particularly crucial to capture the full details of the language learning events, such as learners' prior body of experiences, the sensory richness of the context, and the input and thus influence of caregivers that teach the language. In addition, future research could further investigate the timescale mechanism with respect to hierarchically organized multiple timescales on mathematically more defined tasks, like predicting temporally noisy Lissajous curves with probabilistic transitions (compare Murata et al., 2014) and consider time dilation or time gating, instead of leakage (Chang et al., 2017). Increased understanding and better control of temporal hierarchical composition in neural models, as well as the development of more naturalistic training data and schedules, are promising paths toward models of more human-like language acquisition and learning.

## DATA AVAILABILITY STATEMENT

The datasets generated for this study are available freely via the link provided in the **Appendix**.

## AUTHOR CONTRIBUTIONS

SH developed, implemented, and evaluated the model. SH, YY, THu, and MK collected the data. SH and THi analyzed the model. ZL, CW, and SW helped in writing and revising the paper. All authors contributed to the article and approved the submitted version.

## FUNDING

The authors gratefully acknowledge partial support from the German Research Foundation (DFG) and the National Science Foundation of China (NSFC) under project Crossmodal Learning (TRR-169).

## ACKNOWLEDGMENTS

The authors thank Shreyans Bhansali and Erik Strahl for support in the data collection as well as the NVIDIA Corporation for support with a GPU grant.

**Conflict of Interest:** The authors declare that the research was conducted in the absence of any commercial or financial relationships that could be construed as a potential conflict of interest.

# APPENDIX: EMIL COLLECTION

The Embodied Multi-modal Interaction in Language learning (EMIL) data collection is an ongoing series of data sets for studying human cognitive functions on developmental robots and was first introduced by us during the ICDL-Epirob'2018 workshop on active vision, attention, and learning (Heinrich et al., 2018b). The main motivation is the theory that humans develop cognitive functions from a body-rational perspective. Particularly, infants develop representations through sensorimotor environmental interactions and goal-directed actions (Heinrich and Wermter, 2018). This embodiment plays a major role in modeling cognitive functions from active perception to natural language learning. Using the developmental robotics paradigm, we can investigate specific hypotheses for a range of research questions in-depth, since developmental robotics allows to simulate human development scenarios in a fairly simplified and reproducible way (Cangelosi and Schlesinger, 2015). Thus, data sets that provide low-level multi-modal perception during the environmental interactions are interesting and needed.

With the EMIL data collections, we approach continuous and multi-modal recordings from developmental robot scenarios that specifically focus on robot-object-interaction tasks. Since we aim to utilize resources in tight collaboration with the research community, we propose the first data set on object manipulation in the context of natural language acquisition for closing a gap in current related data sets and fostering discussions on future directions and needs within the community. For the future, we plan to obtain several versions of the EMIL data set with increasing scenario complexity and amount of data. EMIL version 1 is publicly available via:

https://www.inf.uni-hamburg.de/en/inst/ab/wtm/research/corpora.html

## Related Data Sets

In the last years, several labs put considerable efforts into providing data sets on human development scenarios, particularly using the developmental robotics approach. The provided data sets are focusing on different research goals while taking technical limitations into account (see **Table A1**).

As a first example, data sets cover the sensation during human-environment interaction by measuring (mostly adult) humans directly during performing specific tasks, such as the KIT Motion-Language set for descriptions of whole-body poses (Plappert et al., 2016), the Multimodal-HHRI set for personality characterization (Celiktutan et al., 2017), and the EASE set for precise motion capturing (Meier et al., 2018). Secondly, data sets mimic the human perspective by holding objects in front of a perception device, such as a camera, to capture the diverse and complex but general characteristics of an environment setting, e.g., Core50 (Lomonaco and Maltoni, 2017), EMMI (Wang et al., 2017), and HOD-40 (Sun et al., 2018). And thirdly, humanoid robots are employed for establishing a data set, where multiple modalities are recorded in covering human-like action, i.e., including sensorimotor information, such as the MOD165 set (Nakamura and Nagai, 2017) and the Multimodal-HRI set (Azagra et al., 2017), or where multiple modalities are gathered from both robot and human in turn-table actions, like in the HARMONIC data set (Newman et al., 2018).

However, it is usually difficult to capture true continuous multi-modal perception for interaction cases that are supposed to mimic infant experiences or to capture interaction scenarios from human infant learner perspectives. As a consequence, with the EMIL data set collection, we aim to link such continuous multi-modal recordings with body-rationale of a reproducible developmental robot.

## Dataset Characteristics

In this first set, the developmental robot NICO is mimicking an infant that interacts with objects and receives a linguistic label after an interaction. The interaction follows usual interaction schemes of 12–24 month-old infants on toy-like objects.

### Developmental Robot Setup

In developmental robotics, the goal is to study human cognitive functions in conditions of human infants interacting in natural environments (Cangelosi and Schlesinger, 2015). These conditions include *embodied* interaction with natural motor and sensing capabilities of an infant and multi-modal sensations within active perception (Tani, 2016). For our data recording, we developed a child-like humanoid robot and utilize it in scenarios that resemble natural infant environments, such as in playing with objects at a table while acquiring natural language from a caregiver.

### Interactive Robot NICO

Our developmental robot is the Neuro-Inspired COmpanion *NICO* (Kerzel et al., 2017, 2020), created by the Knowledge Technology group of the University of Hamburg. NICO is a research platform that is developed toward research on crossmodal perception, visuomotor learning, and multi-modal human-robot interaction through the embodiment of neurocognitive models. NICO stands about one meter tall with a weight of less than 20 kg. Its proportions follow those of a 3.5-year-old child. Its head is adapted from the open design of the iCub and resembles an abstracted child-like face. Overall, NICO has 30 degrees of freedom that are distributed as follows: each of the legs and arms have six acuted joints. In the arms, three motors in the shoulder area mimic a human ball joint, one motor actuates the elbow, and two motors rotate and flex the hand. Two additional motors in each of NICO's three-fingered, tendon-driven SeedRrobotics hands bend the two linked index fingers and the thumb. The hands allow grasping child-appropriate objects as the tendon-mechanism enables the three-jointed fingers to curl around various shapes without the need for additional control. Finally, two motors enable jaw and pitch motions of the head. For multi-modal sensing, NICO is equipped with two parallel HD RGB cameras and two embedded microphones in its pinnae for stereo auditory perception. Furthermore, the position and current, which is proportional to the applied torque of all motors, can be recorded, which mimics human proprioception. In summary, NICO mimics many of the interaction abilities of a 3.5-year-old child. NICO can handle and

**TABLE A1** | Related multimodal and/or developmental data sets.

| Data set | Modalities | Acquisition | # samples / classes* | Purpose |
|---|---|---|---|---|
| CORe50 (Lomonaco and Maltoni, 2017) | RGB-D vision | Hand-held | 50/10 | Continuous object recognition |
| EASE (Meier et al., 2018) | Vision, audio, motion, EEG, EMG, eye tracker | Human | 100/- | Studying everyday activities for improving robot performance |
| EMMI (Wang et al., 2017) | Vision | Hand-held | 360/12 | Small sample learning; hand object scene interaction |
| EMRE (Krishnaswamy and Pustejovsky, 2019) | Vision, audio | Simulation | 1500/- | Multimodal referring expressions |
| HARMONIC (Newman et al., 2018) | Stereo vision, motion, both robot and human | Turn-table | 480/- | Intention prediction; human mental state modeling; shared autonomy |
| HOD-40 (Sun et al., 2018) | RGB-D vision | Hand-held | 160/40 | Hand-held object recognition; one-shot learning |
| KIT ML (Plappert et al., 2016) | Human motion natural language | Human | 3911/- | Semantic activity representation |
| MHHRI (Celiktutan et al., 2017) | Vision, audio, EDA, skin temp., 3D-accel. | Human | 746/- | Studying personality and engagement |
| MHRI (Azagra et al., 2017) | RGB-D vision, audio | Robot | 300/22 | Incremental object learning from HRI |
| MOD165 (Nakamura and Nagai, 2017) | RGB-D vision, audio, tactile | Robot | 165/- | Studying human-like concepts (ensemble-of-concept model) |

*classes *identify distinct object or action categories, if specified.*

explore physical objects with the imprecision and self-occlusion in a way our infants show.

## Recording

In our experiment, NICO is seated in a child-sized chair at a table, interacting with the right hand and the head facing downwards during the experiment, while a human places a small object on the table at a fixed position (see **Figure 2A**). For EMILv1, a predefined action is carried out on the object: pushing, pulling, lifting it or scooting it across the table. The 30 objects contain toys from an infant environment: balls, toy cars, sponges and tissues, fruits, small animals, and toy bricks, of which some differ in softness during squeezing, weight, size, and color. During the robot's actions, a continuous multi-modal recording encompasses continuous streams of visual information from the left and right robot camera as well as from the external experimenter, stereo audio information from microphones in the robot's head, and proprioceptive information from the robot's body, specifically position and current from eight motors (for an example compare the input streams in **Figure 3**). Finally, the experimenter provides a linguistic label.

## Preprocessing

To provide the data in suitable formats for various research questions, we added preprocessed versions of the raw data as follows. For the auditory signals, we added streams of Mel-Frequency Cepstral Coefficients (MFCC) transformation with 13 dimensions, a frame size of 33 ms, and input window 60 ms. Using filters with Mel-scale is considered biologically-inspired as this mimics the humans' perception of frequencies and the sensitivity of the cochlea, which can be seen as kind of a Fourier transformation of auditory signals. The frame size is motivated in the technical characteristics of the motor sensors and the cameras and is supposed to allow for obtaining an aligned frame rate. The MFCCs overlap with 50% because the Fourier transformation creates border effects, which the window size of 60 ms is acceptable since we mostly record environmental noise. Because of the volatile nature of the position and current sensors in the motor we produced smoothed sensorimotor streams based on 3, 5, 7, and 9 measurement points. We also normalized all sensorimotor streams w.r.t. the minimal and maximal position and current values per joint. For the visual streams, we offer compressed videos with a cropped field of view (e.g., only the table or only the interesting part of the table) for convenience.

## Labeling for Object Tracking

For supporting research questions related to object tracking we added a complete ground-truth labeling for all visual streams from the perspective of NICO's right eye. The object labeling describes the position of the interacted object in all frames with accurate bounding boxes despite strong transformations and occlusions.

## Labeling for Language Learning

All interactions are labeled textually with words describing the action and the object type, as well as particularly deviating object characteristics (color, weight, softness, size). Depending on the research question with relation to natural language processing, different textual utterances or descriptions can be generated. For instance, EMILv1 is provided with labels in the form of holo-phrases with up to four words as well as additional labels containing synonyms for the actions and object characteristics (compare section 2.3).

## Impact and Research Opportunities

Our continuous, multi-modal, and particularly body-rational data allows for studying a large range of algorithms on fundamental classification or prediction tasks. This includes object recognition and tracking, action recognition, and question answering. Moreover, the data set is aimed at research on a range of state-of-the-art research topics.

### Active Perception

The different actions and objects allow to build up a training scheme within a model by selecting to experience a certain interaction because the model estimates that this provides the highest information gain or reduces uncertainty. In humans, we find the tendencies that a perception choice or a specific action is voluntary (Oudeyer, 2018). Thus, the data set is suited for developing models that aim to explain how the sensory input gathered from an object with different, multi-modal sensors changes based on the robot's actions.

### Imitation Learning

Robotic visuomotor learning via interaction with the environment often requires a large amount of training data and, therefore, physical interactions (Lillicrap et al., 2016), which are not feasible for most robotic platforms. However, one way of accelerating the learning process is to utilize demonstrations to speed up the initial learning phase. While the demonstrations are usually provided by humans (Gupta et al., 2016), the precise motor data in the EMIL data set can be utilized for this purpose as well with the added benefit that this data is free of artifacts or noise from an external recording setup.

### Cross-Modal Representation Learning

Since the different recorded modalities include information about the same object and interaction quite differently, the data set is suited to study algorithms on multi-modal and cross-channel representation learning. For some objects and actions the data contains salient features in a certain modality, while for others, all modalities are necessary for disambiguation. This allows studying mechanisms on sensor fusion, superadditivity, and hierarchical composition in addition to embodied representation formation on the cortex-level (Bauer et al., 2015).

### Developmental Language Acquisition

A research question related to representation learning is natural language acquisition since representations for language production and language perception in the human brain seem to form embodied and cross-modally integrated (Cangelosi and Schlesinger, 2015; Heinrich and Wermter, 2018). The data set is therefore particularly suited for research on the grounding of language in sensorimotor perception because the recording diligently followed the developmental robot approach (Lyon et al., 2016). Mechanisms for representation formation and bidirectional hierarchical composition and decomposition can get tested in the biologically plausible setting.

As a second step, this allows extending this data set by much larger parts of abstract and ungrounded linguistic input, in a fashion that parents would provide verbally or with the aid of a storybook to their infant (Heinrich et al., 2016). Here, language acquisition models can get studied for how they integrate additional knowledge into their grounded representations, but also how a teaching application can provide suitable teaching content.

### Lifelong Learning

The data set is suited to provide evaluation data for (neural) lifelong learning approaches (Parisi et al., 2018). An initial subset of the training data can be selected that is limited to a few types of objects, actions or just a low number of samples. Over the course of time, lifelong learning experiences can be simulated by adding more and more parts of the data-set to the learning.